\documentclass[10pt,twocolumn,letterpaper]{article}

\usepackage{cvpr}      % To produce the REVIEW version
\usepackage{amsmath,amssymb,amsfonts}  % 数学
\usepackage{caption}                   % 调整 caption 样式

\usepackage[ruled,vlined]{algorithm2e}
\SetKwInput{KwIn}{Input}
\SetKwInput{KwOut}{Output}

\definecolor{cvprblue}{rgb}{0.21,0.49,0.74}
\usepackage[pagebackref,breaklinks,colorlinks,allcolors=cvprblue]{hyperref}

\def\confName{CVPR}
\def\confYear{2026}

\title{RefAerial: A Benchmark and Approach for Referring Detection in Aerial Images}

\author{Guyue Hu$^{1,3,4,5}$, Hao Song$^{1,3,4,5}$, Yuxing Tong$^{1,3,4,5}$, Duzhi Yuan$^{1}$, Dengdi Sun$^{1,4,5}$,Aihua Zheng $^{1,3,4,5}$\thanks{Corresponding author: {\tt\small ahzheng214@foxmail.com}},\\ Chenglong Li$^{1,3,4,5}$, \
Jin Tang$^{2,5}$\\
 $^1$School of Artificial Intelligence, Anhui University\\
 $^2$School of Computer Science and Technology, Anhui University, Hefei, China\\
 $^3$State Key Laboratory of Opto-Electronic Information Acquisition and Protection Technology, Anhui\\ University, Hefei, China  \\
 $^4$Anhui Provincial Key Laboratory of Security Artificial Intelligence, Anhui
University, Hefei, China\\
 $^5$Anhui Provincial Key Laboratory of Multimodal Cognitive Computation, Anhui
University, Hefei, China\\
}

\begin{document}
\maketitle
\begin{abstract}
Referring detection refers to locate the target referred by natural languages, which has recently attracted growing research interests.\hspace{0.5em}However, existing datasets are limited to ground images with large object centered in relative small scenes.\hspace{0.5em}This paper introduces a large-scale challenging dataset for referring detection in aerial images, termed as RefAerial.\hspace{0.5em}It distinguishes from conventional ground referring detection datasets by 4 characteristics:\hspace{0.5em}(1) \textit{low but diverse object-to-scene ratios}, (2) \textit{numerous targets and distractors}, (3) \textit{complex and fine-grained referring descriptions}, (4) \textit{diverse and broad scenes in the aerial view}.\hspace{0.5em}We also develop a human-in-the-loop referring expansion and annotation engine (REA-Engine) for efficient semi-automated referring pair annotation.\hspace{0.5em}Besides, we observe that existing ground referring detection approaches exhibiting serious performance degradation on our aerial dataset since the intrinsic scale variety issue within or across aerial images.\hspace{0.5em}Therefore, we further propose a novel scale-comprehensive and sensitive (SCS) framework for referring detection in aerial images. It consists of a mixture-of-granularity (MoG) attention and a two-stage comprehensive-to-sensitive (CtS) decoding strategy. Specifically, the mixture-of-granularity attention is developed for scale-comprehensive target understanding. In addition, the two-stage comprehensive-to-sensitive decoding strategy is designed for coarse-to-fine referring target decoding. Eventually, the proposed SCS framework achieves remarkable performance on our aerial referring detection dataset and even promising performance boost on conventional ground referring detection datasets.
\end{abstract}    
\section{Introduction}
\begin{figure}[!t]  % 使用figure环境可以为图片添加标题，并自动管理图片位置
    \centering
    \includegraphics[page=1, width=0.85\linewidth, keepaspectratio]{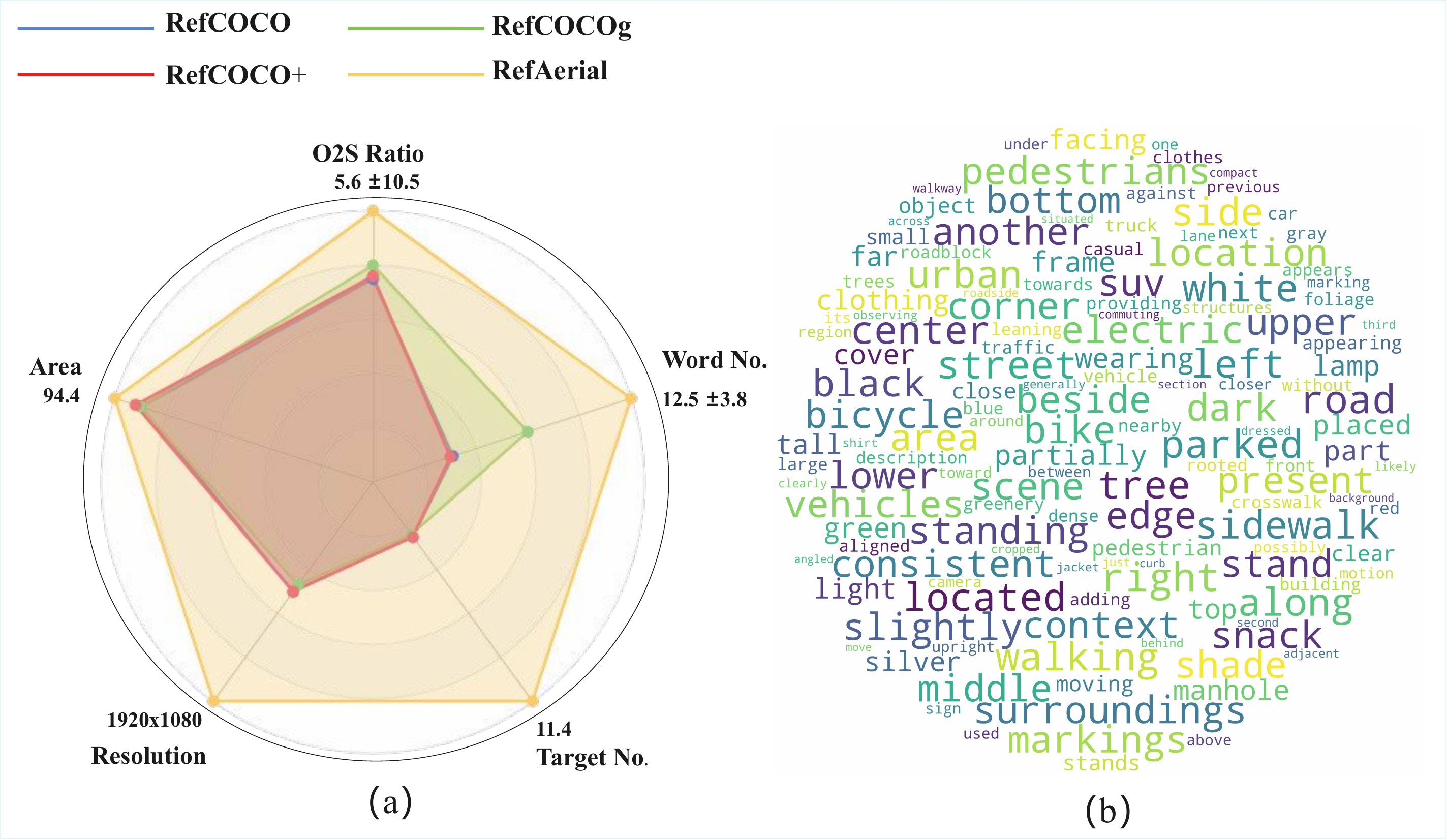} 
    \caption{(a) Comparing ground referring detection datasets with our RefAerial dataset. Non-target Area (opposite object-to-scene ratio) indicates the proportion of non-target regions in aerial images. Target No.: refers to the average number of referring objects per image. Word No.: the number of word tokens per expression. Resolution represents the average resolution of images in datasets. (b) The word cloud of referring descriptions in the RefAerial dataset. The words related to fine-grained attributes, objects, actions, positions, and colors appear frequently.} 
    %In the RefAerial dataset, words related to spatial positions or directions occur with high frequency and are used to describe the relative positions of objects.}  % 图片标题
    \label{fig:1}  % 可选的标签，便于在文档中引用
\end{figure}
\label{sec:intro}
\begin{figure*}[!t]  % 使用 figure* 环境横跨双栏，!t 强制尽量置顶
    \centering
    \includegraphics[page=3, width=0.8\textwidth, keepaspectratio]{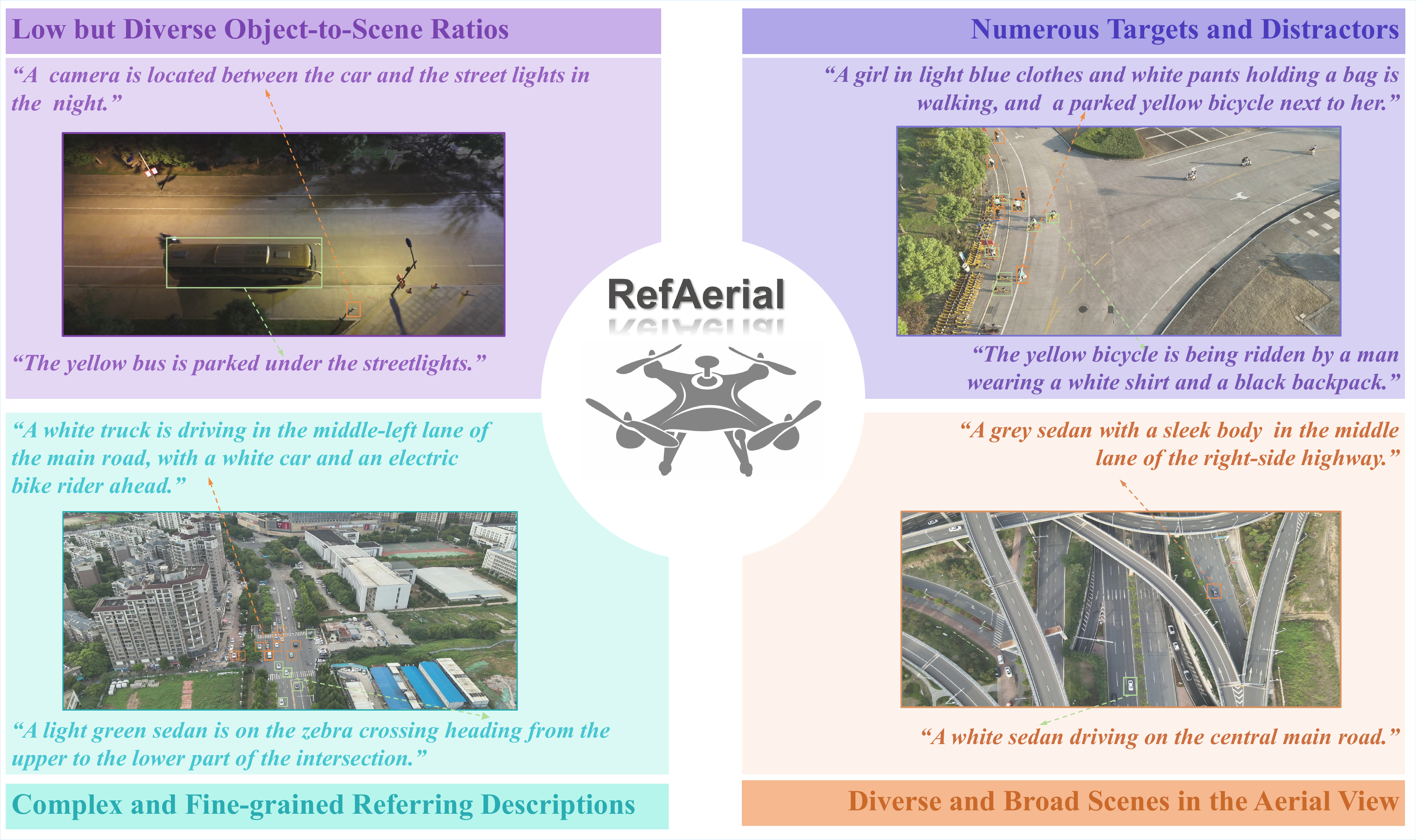}
    \caption{Illustration of challenges and samples in the RefAerial dataset.}
    \label{fig:2}
\end{figure*}
Referring detection (a.k.a visual grounding) aims at locating the target regions in an image referred by a given descriptive language expression.\hspace{0.5em}It has wide applications, such as human-computer interaction \cite{CHI3, CHI2}, intelligent perception \cite{IP1, 6G}, and vision-and-language navigation \cite{song2025towards, zhuvision, li2024language}. However, existing datasets for referring detection (such as RefCOCO \cite{refcoco} and RefCOCOg \cite{refcocog}) mainly focus on ground images, in which the referring targets usually are centered in small ground scenes and also relatively large with high object-to-scene ratios.\hspace{0.5em}In the past few years, the quick popularity of low-altitude economy and drone-based imaging devices has driven urgent demands of studying referring detection in broad aerial-view scenes, since its wide and significant applications in emergency rescue, agricultural monitoring, ecological protection, anti-drone confrontation, etc.

\begin{table}[tpb] %h表示三线表在当前位置插入
\setlength{\abovecaptionskip}{0.05cm} %设置三线表标题与第一条线间距
\centering
\caption{Comparing our RefAerial with existing ground referring detection datasets. The RefAerial significantly distinguishes from them regarding object-to-scene ratios, target numbers, description complexity, bounding-box numbers, and scene scenarios. O2S Ratio (\text{\%}): object-to-scene ratio in the form of Mean (Standard Deviation). Word No.: number of words per expression in the form of Mean (Standard Deviation). Target No.: average number of annotated referring targets per image. Bbox No.: total number of bounding boxes in the whole dataset. Resolution represents the average resolution of images in the dataset.}
%表头文本加黑，但不加黑Table 1.字样，引入包即可：\usepackage[labelfont=bf]{caption}
 %设置三线表线条颜色：黑色
\setlength{\tabcolsep}{1mm}
\small % 默认小一号字体
\begin{tabular*}{\hsize}{@{\extracolsep{\fill}}l | c c c | c} %{\hsize}使三线表自适应宽度，c表示文本居中
  \hline
  \hline
  \textbf{Items} & \textbf{RefCOCO} & \textbf{RefCOCOg} & \textbf{RefCOCO+} & \textbf{RefAerial}\\
  \hline 
  O2S Ratio &11.4 (8.5) & 11.0 (9.4)  & 11.1 (8.6) &\textbf{5.6} (\textbf{10.5}) \\
  Word No. & 3.61  (2.0)  & 8.43 (3.5) & 3.53 (2.1)  & \textbf{12.5} (\textbf{3.8})  \\
  Target No. & 2.5  & 1.9 & 2.5 & \textbf{11.4}\\
  %Bbox ~No. & 51,324 & 54,822 & 49,856 & \textbf{104,517}  \\
  Bbox ~No. & 51K & 55K & 50K & \textbf{115K}  \\
  Resolution & 592×485  &  583×480 & 592×485  & \textbf{1920x1080}\\
  \hline
  \hline
\end{tabular*}
\label{tab:1}
\end{table}

To activate and advance the research and application of referring detection in aerial images, we collect and annotate a large-scale challenging datasets for referring aerial image detection, termed as RefAerial. It distinguishes from classical ground referring detection datasets by 4 remarkable characteristics (as shown in Fig. \ref{fig:2}).
(1) \textbf{Low but diverse object-to-scene ratios:} since the aerial devices inherently have broader scope compared to ground devices, the referring object in aerial images typically occupies small portion of pixels in the overall scene. In addition, diverse scales of objects are commonly contained within and across aerial images (see Table \ref{tab:1} and Fig. \ref{fig:1}a). Thus, the low but diverse object-to-scene ratio significantly increases the difficulty of object discrimination, spatial localization, and boundary estimation for referring detection in aerial images.
(2) \textbf{Numerous targets and distractors:} an aerial image typically contains numerous potential target categories and each referring target in one specific semantic category may also be accompanied with numerous similar distractors nearby (such as the girl and bicycle in Fig.\hspace{0.5em}\ref{fig:2}), please see Table \ref{tab:1} for more details. Thus, it is very challenging to recognize and discriminate the referring target from other distractors with contextual information. (3) \textbf{Complex and fine-grained referring descriptions:} the descriptions in mainstream ground referring detection datasets are usually short and coarse-grained, while our RefAerial dataset covers complex (long and diverse) and fine-grained referring descriptions (see Fig. \ref{fig:1}b and Table \ref{tab:1}). Its descriptions have an average length of 12.5 words, and short and long descriptions are nearly balanced. Thus, the complex and fine-grained referring descriptions substantially increase the linguistic complexity. (4) \textbf{Diverse and broad scenes in the aerial view:} in contrast to small and centralized ground scenes, our RefAerial are collected from broad aerial scenes with diverse environmental conditions regarding altitude, view, weather, area, etc.

Besides, we observe that existing ground referring detection approaches face serious performance degradation on our aerial referring detection dataset since the intrinsic scale variety issue within or across aerial images (as shown in Table \ref{tab:1}). To address such issue, we further propose a novel scale-comprehensive and sensitive (SCS) framework characterized with a mixture-of-granularity (MoG) attention and a two-stage comprehensive-to-sensitive (CtS) decoding strategy. Specifically, the mixture-of-granularity (MoG) attention is developed to scale-comprehensively understand potential targets in each aerial image. The comprehensive-to-sensitive decoding (CtS) strategy is designed to decode referring targets in a progressive coarse-to-fine manner.

Overall, the main contributions of this work are 4 folds:
\begin{itemize}
 \item Responding to low-altitude economy, we extend referring detection from conventional ground images to emerging aerial images and build a large-scale dataset  RefAerial for referring aerial image detection.
 \item  We develop a human-in-the-loop referring expansion and annotation engine (REA-Engine), which realizes efficient and semi-automated referring pair annotation.
 \item We propose a novel scale-comprehensive and sensitive framework (SCS) for referring aerial image detection, which effectively tackles the remarkable scale variety issue of referring objects within and across aerial images.
 \item The proposed SCS framework realizes remarkable performance on our aerial dataset and even promising performance boost on ground datasets, demonstrating the effectiveness and superiority of our approach.
\end{itemize}

\section{Related Work}
\subsection{Referring Detection Datasets}
With the rapid advancement of visual technologies, an increasing number of referring detection (a.k.a visual grounding) datasets have been proposed, such as RefCOCO \cite{refcoco}, RefCOCO+ \cite{refcoco}, RefCOCOg \cite{refcocog}, and Flickr30k \cite{flickr30k}. Specifically, RefCOCO emphasizes interactive short phrases, RefCOCO+ restricts the use of spatial terms to increase semantic reasoning difficulty, and RefCOCOg focuses on long and context-rich expressions. These datasets have collectively driven significant technical advances of referring detection, such as multimodal modeling, cross-modal alignment, and language-guided localization. Despite serving as a standard evaluation benchmark, the RefCOCO series of datasets are inherently expanded from COCO dataset, in which the referring targets usually are centered in small ground scenes and also relatively large. In contrast, the benchmark and approach for referring detection in broad aerial-view scenes remains underdeveloped. 
%which tend to be relatively "clean"—with fewer occlusions, less inter-class interference, and limited target ambiguity—thereby constraining the generalization capabilities of models in more complex and realistic environments.
%随着视觉技术的快速发展，越来越多的指称检测（又称视觉接地）数据集被提出，例如 RefCOCO（Yu et al., 2016）、RefCOCO+（Yu et al., 2016）、RefCOCOg（Mao et al., 2016）以及 Flickr30k（Plummer et al., 2015）。现有研究在这些包含相对简单视觉环境的数据集上取得了显著成果。例如，COCO 数据集涵盖 80 个常见日常物体类别，图像来源丰富，场景类型多样，具有很强的代表性和通用性。RefCOCO 系列数据集以 COCO 数据集 (\cite{XXX}) 中的地面自然图像为基础，进一步融入人工标注的指称表达，从而将多物体图像与指称检测任务连接起来。这些表情通常针对单个目标进行标注，综合考虑空间位置、外观、语义等特征，从而明确地映射语言描述和视觉对象。具体而言，RefCOCO 强调可交互的短语，RefCOCO+ 限制空间词的使用以增加语义推理难度，而 RefCOCOg 则侧重于长且上下文丰富的表情。这些数据集共同推动了指称检测的重大技术进步，例如多模态建模、跨模态对齐和语言引导定位。尽管 RefCOCO 系列数据集作为标准的评估基准，但它本质上是从 COCO 数据集扩展而来的，COCO 数据集中的指称目标通常集中在较小的地面场景中，并且相对较大。相比之下，用于鸟瞰场景中指称检测的基准和方法仍然不够完善。

\subsection{Referring Detection Methods} 
Referring expression comprehension methods are typically categorized into two paradigms: two-stage methods and one-stage methods. Two-stage approaches \cite{CMN, pa, mn, nms}  first generate region proposals with an object detector and then match these candidates with the given expression. These methods often yield high localization accuracy but suffer from limited inference efficiency, which hinders their applicability in real-time scenarios. In contrast, one-stage methods \cite{fast, rccfl, sub} jointly encode visual and linguistic inputs in an end-to-end manner to directly predict the referred object, offering superior efficiency and broader potential for real-world deployment. Recently, Transformer-based architectures \cite{qrnet, vltvg, polyformer, scanformer} have been widely adopted for multimodal fusion for referring detection, yielding significant performance improvement. Additionally, several works fine-tuned large-scale pre-trained Vision-Language Models (VLMs) \cite{qwen, hivg, dual, approximated}, leveraging their strong semantic priors to further enhance grounding performance. Although these high success of referring detection approaches in ground images, these face serious performance degradation on our aerial referring detection dataset, thus encouraging the exploring of more robust and adaptable methods for referring detection in aerial images.

\section{RefAerial Dataset}
In this section, we introduce how to collect, and then expand and annotate the aerial RefAerial dataset via the proposed semi-automated human-in-the-loop referring expansion and annotation engine (REA-Engine) in detail. Then, we highlight its statistics and characteristics compared to existing ground referring detection datasets.

%for  referring detection in aerial images in detail. %Then, we primarily highlight the statistics and characteristics compared to existing ground referring detection datasets.

\begin{figure*}[!t]  % 使用 figure* 环境横跨双栏，!t 强制尽量置顶
    \centering
    \includegraphics[page=1, width=0.87\linewidth]{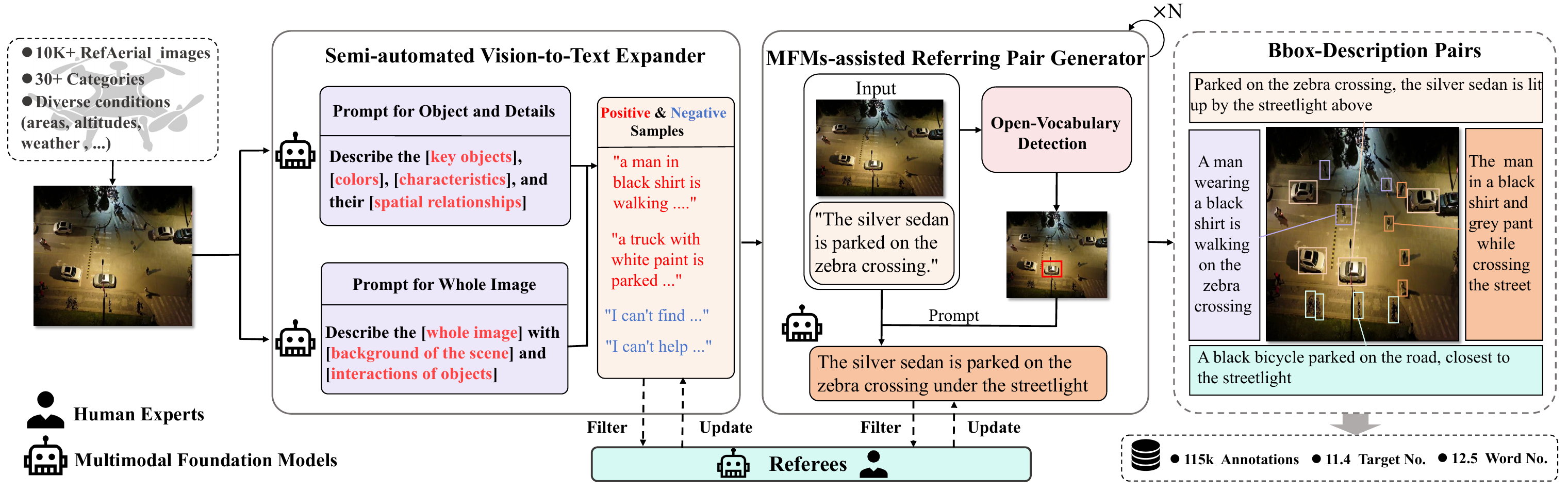}
    \caption{Overview of the referring expansion and annotation engine (REA-Engine). It is a semi-automated human-in-the-loop annotation framework that integrates large-scale multimodal foundation models (MFMs) with human experts to generate high-quality referring pair annotation for aerial images.}
    \label{fig:3}
\end{figure*}

\subsection{Aerial Image Collection}
%We first collect abundant aerial images with high-resolution (1920x1080) drone devices across diverse conditions in the wild. As shown in Table \ref{tab:11}, It contains over 30 categories of aerial view targets (such as people, car, tree, etc.) from diverse environments, including areas (such as urban and suburb), locations (such as university, park, crossroad, lake, etc.), altitudes (30m, 90m, 150m), view angles ranging from 0$^{\circ}$ to 90$^{\circ}$, covering steep-oblique to near-nadir perspectives, lighting conditions (such as day and night), weather conditions (clear, cloudy, light rain), scale, and so on. Thus, the aerial images in RefAerial contain enough representativeness and diversity in real-world aerial referring detection.

We first collect abundant aerial images with high-resolution (1920×1080) drone devices under diverse real-world conditions. The data are collected in diverse outdoor environments to cover various scene complexities, viewpoints, and object types. As shown in Table~\ref{tab:11}, RefAerial dataset contains more than 30 categories of aerial-view targets (such as people, cars, trees, buildings, etc.), covering multiple types of environments and geographical contexts. It include different area types (urban and suburban), representative locations (university, park, crossroad, lake, residential block, etc.), and multiple flight altitudes (30m, 90m, 150~m). The view angles ranging from 0$^{\circ}$ to 90$^{\circ}$,  covering steep-oblique to near-nadir perspectives, which introduces large appearance and scale variations. Moreover, the dataset encompasses diverse illumination (daytime, overcast, nighttime) and weather conditions (clear, cloudy, light rain). Thus, the aerial images in RefAerial contain enough representativeness and diversity for real-world aerial referring detection.

%The RefAerial dataset is a newly captured collection of drone-view images, consisting of 9300 high-resolution images. It covers over 30 different object categories and includes a total of 104,517 labeled objects. The source images are captured across multiple scenarios, lighting conditions, and flying altitudes, providing rich visual data for various object detection tasks. Each image in RefAerial is carefully annotated with object categories and bounding box coordinates, making it an ideal resource for training and evaluating machine learning models focused on aerial object detection.

\subsection{Referring Expansion and Annotation Engine}
Annotating the visual bounding-box and textual description pair of referring targets is much complicated and expensive. Beyond manual annotation, we develop an advanced referring expansion and annotation engine (REA-Engine) that leverages the capabilities of powerful large multimodal foundation models (MFMs) for collaborative enhancement. As a result, REA-Engine enables semi-automated human-in-the-loop annotation workflows to produce fine-grained and reliable aerial referring detection datasets. As shown in Fig. \ref{fig:3}, REA-Engine consists of two modules (including the Vision-to-Text Expander and MFMs-assisted Referring Pair Generator) guided by feedback from referees.

%Owing to the intrinsic complexity of Referring Detection tasks, the construction of training datasets demands extensive annotations across both textual and visual modalities. Beyond manual labeling, we develop an advanced data engine that leverages the capabilities of powerful foundation models for collaborative augmentation, enabling human-in-the-loop annotation workflows to produce high-quality and reliable visual grounding datasets.

\subsubsection{Semi-automated Vision-to-Text Expander}
We first develop a vision-to-text expander (V2T expander) to efficiently expand aerial images with detailed language description containing global scene, salient object, fine-grained details for referring pair generating. The V2T expander is a semi-automated human-in-the-loop workflow that leverages the powerful multimodal comprehend capabilities in off-the-shelf large MFMs (such as GPT4 \cite{gpt} and CogVLM \cite{cogvlm}) in conjunction with the guidance from human experts. As shown in Fig. \ref{fig:3}, the V2T expander consists of 3 levels of semantic prompts and receives feedback from referees. Specifically, the global scene prompt elicits holistic scene-level description (such as contextual background), the salient object prompt guide the description of key entities, and the fine-grained prompt encompasses detailed attributes and relations (such as color, position, interactions). Furthermore, the referees empowered by MFMs and human experts operate in a closed feedback loop with follow-up MFMs-assisted referring pair generator and iterative human validation, facilitating continual improvement in both granularity, precision, and coverage of language descriptions.

%\textbf {Modality Expansion Annotator.}
%To address the challenge of scalable and semantically rich multimodal annotation, we propose a Modality Expansion Annotator, a human-in-the-loop annotation module that leverages the expressive capabilities of large language models (LLM)\cite{gpt,cogvlm,shikra} in conjunction with expert guidance. Specifically, our annotator supports dual-level semantic prompting: salient object prompts guide the description of key entities and their spatial relationships (e.g., “in front of”, “next to”), while global image prompts elicit holistic scene-level descriptions, encompassing contextual background and object interactions. To ensure consistency and annotation quality, we integrate a referee model that automatically refines and filters human-generated text, aligned with high-quality exemplar patterns. This process enables the construction of fine-grained, open-vocabulary image-text-bounding box triplets, forming a crucial foundation for robust visual grounding and multimodal understanding. Furthermore, the annotator operates in a closed feedback loop with model-assisted generation and iterative human validation, facilitating continuous improvement in both coverage and precision of annotations.

\subsubsection{MFMs-assisted Referring Pair Generator}
We further develop a semi-automated MFMs-assisted referring pair generator (MaRP generator) to generate and refine the referring pair of \{Bbox, description\} from above preliminary descriptions. Specifically, we finetune a large foundation segmentation model (\ie~SAM \cite{SAM}) as internal object detector to recognize and localize object entities referred to by coarse language expressions. Then, the image with object bounding box (Bbox) annotation will feed into a large MFMs again to generate more fine-grained and accurate referring descriptions, the external knowledge contained in general MFMs also enriches the semantic diversity of the detection descriptions. Finally, the MaRP generator generated pair candidates will be forwarded into the referees, where the MFMs and human experts further collaboratively refine to obtain the final referring pairs. In details, the referees conduct consistency validation, redundancy removal, and semantic correction to guarantee each referring pair is fine-grained, accurate, visible, and information-dense. During this process, the MFMs automatically filter and score image–text pairs, merging near-duplicate descriptions into unified candidates with associated confidence scores. Human experts review the high-confidence samples, providing corrections and quality judgments. Their feedback is then used to retrain and calibrate the MFMs, forming a closed-loop refinement cycle that typically converges within 3–5 iterations.  The MaRP generator not only significantly reduces the burden of manual annotation, but also strengthens the capability of object perception and language understanding under complex scenarios, which lays a solid foundation for high-quality Bbox-description pairs.

\begin{figure*}[!t]  % 使用 figure* 环境横跨双栏，!t 强制尽量置顶
    \centering
    \includegraphics[page=2,width=0.8\textwidth]{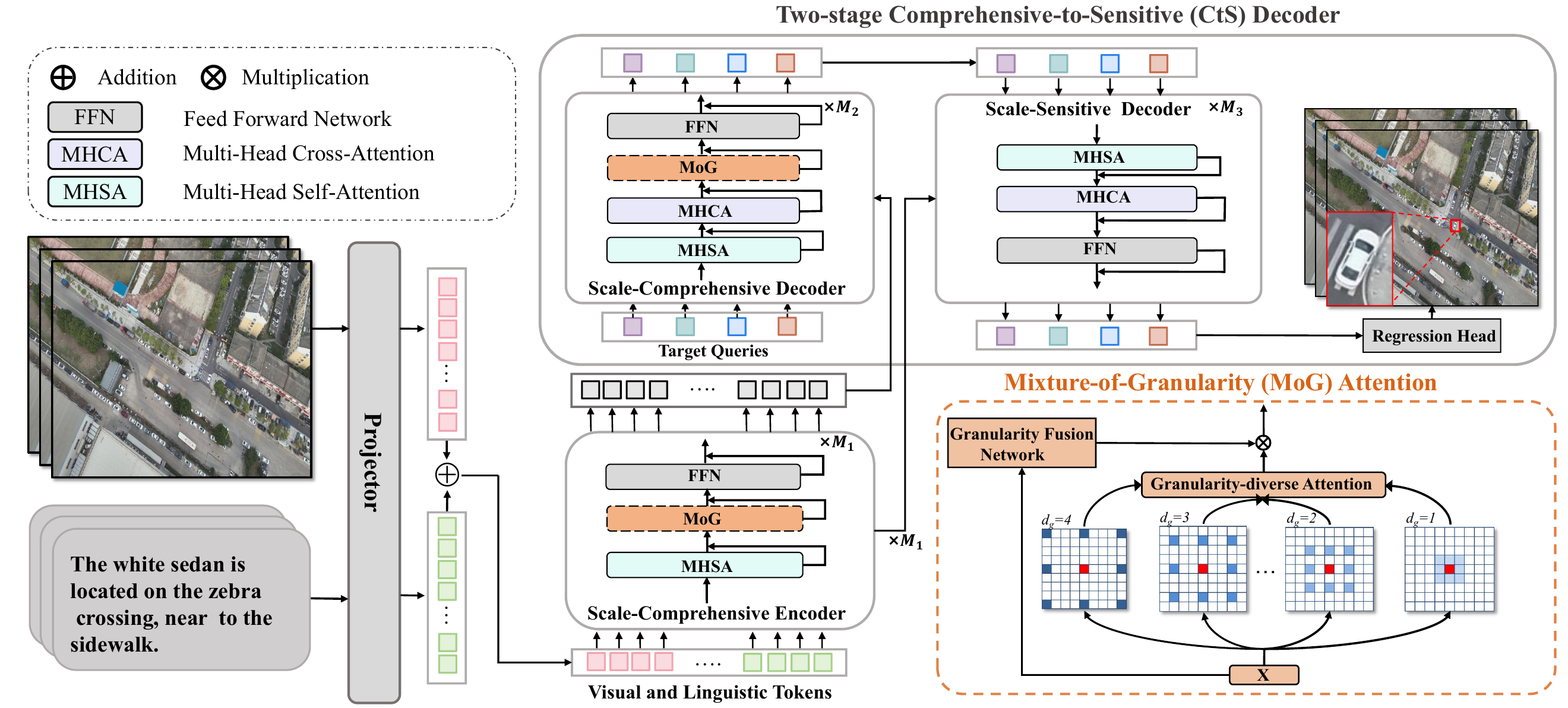}
    \caption{The overall pipeline of the proposed scale-comprehensive and sensitive (SCS) framework. It is characterized with a mixture-of-granularity (MoG) attention and a two-stage comprehensive-to-sensitive (CtS) decoding strategy. In detail, the SCS framework is in form of the DETR-style architecture \cite{detr} consisting of a token projector, a scale-comprehensive encoder (SCE), a scale-comprehensive decoder (SCD), a scale-sensitive decoder (SSD), and a regression head.}
    
    %The architecture of SCS framework.The proposed SCS framework introduces a Mixture-of-Granularity (MoG) attention mechanism and a two-stage Comprehensive-to-Sensitive (CtS) decoder to effectively address the challenges of diverse object scales and complex distractors in aerial images. The model first encodes image and text features separately, then uses the MoG module to extract multi-scale features. A coarse target location is obtained through the first-stage decoder, which is further refined by the second-stage decoder, ultimately achieving accurate target localization.
    \label{fig:4}
\end{figure*}

\renewcommand{\thetable}{2}
\begin{table}[t]
\centering
\captionsetup{aboveskip=1pt,belowskip=2pt}
\caption{Condition statistics of aerial images in RefAerial dataset.}
\large
\setlength{\tabcolsep}{2.5mm} % 调整列间距
\resizebox{\linewidth}{!}{%
\begin{tabular}{l|ccc|ccc} % 保持竖线位置
\hline
\hline
\textbf{Condition} & \multicolumn{3}{c|}{\textbf{Category}} & \multicolumn{3}{c}{\textbf{Statistics}} \\  % 这个地方的竖线需要调整
\hline
Flight Altitude (m) & 30 & 90 & 150 & 32\% & 40\% & 28\% \\
Camera Angle (°) & 0--30 & 30--60 & 60--90 & 25\% & 57\% & 18\% \\
Illumination & Day & Overcast & Night & 61\% & 21\% & 18\% \\
Weather & Clear & Cloudy & Light rain & 65\% & 22\% & 13\% \\
Scene Type  &  & 25 &  &   & /  &   \\ 
Target Category   &  & 32  &  & & / &  \\ 
\hline
\hline
\end{tabular}%
}
\label{tab:11}
\end{table}

\subsection{Dataset Details} 
In this section, we provide statistics of our RefAerial and compare it with existing ground referring detection datasets. 

The RefAerial consists of 10056 high-resolution aerial images (1920x1080) and 114,517 (115K) Bbox-description pairs for referring detection. As shown in Table \ref{tab:1} and Fig. \ref{fig:1}a, each aerial image contains 11.4 referring targets (Target No.) on average, while ground images in existing datasets only contain 1-3 targets per image, making RefAerial inevitably challenging in numerous targets and distractors. In addition, our RefAerial has very low (Mean: 5.6\text{\%}) but diverse (STD: 10.5\text{\%}) object-to-scene ratios, while the referring targets in existing ground referring detection datasets are usually centered and have high object-to-scene ratios (average larger than 10\%). Meanwhile, our RefAerial is annotated with more complex and fine-grained referring descriptions with 12.5 words per description on average. The word cloud analysis of referring descriptions in Fig. \ref{fig:1}b further highlights the most frequent mentioned words are relevant to fine-grained elements, such as attributes, actions, objects, positions, colors, and so on. Eventually, the visualization of these comparisons is shown in Fig. \ref{fig:1}a for intuitively understanding.

%word cloud analysis s (as shown in Fig.2) highlights the frequent mentions of target types, colors, and spatial positions. These annotations place greater emphasis on relative location rather than the absolute position of the target. This rich annotation style is crucial for the task and also presents greater challenges for the models.

%Fig.3 compares the RefAerial dataset with existing foundational visual grounding datasets across key aspects, including image resolution, average number of words in textual queries, average number of objects per image, information entropy of the queries, and the proportion of irrelevant regions. In all these aspects, the RefAerial dataset exhibits a higher level of challenge, highlighting its complexity and suitability for more advanced visual grounding tasks.

\section{Method}
\subsection{Overview}

To alleviate serious performance degradation (see Table \ref{tab:4}) caused by the intrinsic scale variety issue within or across aerial images, we propose a novel scale-comprehensive and sensitive (SCS) framework tailored for referring detection in aerial images. As shown in Fig. \ref{fig:4}, it is characterized with a mixture-of-granularity (MoG) attention and a two-stage comprehensive-to-sensitive (CtS) decoding strategy.\hspace{0.5em}The MoG attention is developed for scale-comprehensively understanding potential targets in aerial images, while the two-stage CtS strategy is designed to coarse-to-finely decode referring targets. We introduce the preliminary pipeline of SCS framework in the following.

As shown in Fig.\hspace{0.5em}\ref{fig:4}, we first adopt ResNet \cite{resnet} and BERT \cite{bert} as projectors to embed image-description pairs as visual and linguistic tokens, respectively. Then, their concatenation is fed into classical multimodal DETR-style pipelines (such as SegVG \cite{segvg}) to go through multimodal scale-comprehensive target encoding and two-stage comprehensive-to-sensitive target decoding. Eventually, the referring target will be well perceived and localized in the corresponding aerial image.

\subsection{Mixture-of-Granularity Attention}
In contrast to conventional large centralized targets in ground referring images, the potential referring targets in aerial images usually exhibit significant scale variation issue within or across aerial images since the intrinsic characteristics of drone devices (such as the altitude-varying and broad scene). It is challenging for traditional scale-unaware multi-head attentions in Transformer to understand these scale-varying referring targets within or across aerial images, thus we developed a novel mixture-of-granularity (MoG) attention to realize scale-comprehensive encoding and decoding.  

%In aerial-view images captured from a UAV perspective, targets often exhibit significant scale variations and complex spatial distributions, making it challenging for traditional self-attention mechanisms to balance computational efficiency and modeling capacity. To address this issue, we propose a Mixture-of-Granularity Attention (MoG Attention) mechanism, which aims to enhance the model’s ability to perceive multi-scale targets and improve its generalization performance in complex scenes.

Specifically, the proposed MoG attention module contains multiple sub-attentions with varying receptive fields (dilation rates) instead of standard token-wise attention to encourage multi-granularity perception in scale-varying aerial images. As shown in Fig. \ref{fig:4}, the fine-grained branches with low dilation rates tend to focus more on local details which are more suitable for capturing small targets and detail boundary. In contrast, the coarse-grained branches emphasize large dilation rates tend to emphasize more non-local information which is useful for capturing large-scale structures and non-local relationships. 

Formally, given the representation sequence \( \mathbf{X} \in \mathbb{R}^{B \times N \times D} \), where B, N, D  are the batch size, sequence length, embedding dimension, respectively. It first undergoes the same process of the standard attention to obtain query \textbf{Q}, key \textbf{K}, value \textbf{V}, and corresponding attention weight  $\mathbb{A}$, \ie 
\begin{equation}
\mathbf{Q}, \mathbf{K}, \mathbf{V} = \text{linear}_{\{q,k,v\}}(\mathbf{X})
\end{equation}
\begin{equation}
\mathbf{A} = \frac{\mathbf{QK}^\top}{\sqrt{d_k}}
\end{equation}
where  $\text{Linear}_{\{q, k, v\}}(\cdot)$  is the linear projection function for query, key, and value. Then, as shown in Fig. \ref{fig:4}, each granularity of sub-attention is equipped with a distinct dilation rate \( d_g \) to sparsify the receptive field of the corresponding attention branch. In detail, we define a binary mask \( \mathbf{M}^{(g)}  \) for each granularity \( g \) which only retains token pairs satisfying distance dilation rate  \( d_g \), \ie
\begin{equation}
\mathbf{M}_{i,j}^{(g)} = \mathbb{I}(\bmod (|i - j|, d_e) = 0)
\end{equation}
where $i, j$ are token position indices, $\mathrm{mod}$ denotes the modulo operation, and $\mathbb{I}(\cdot)$ is the indicator function. Thereafter, the new attention weight $\mathbf{A}'$ is defined as follows,
\begin{equation}
\mathbf{A}' = \mathbf{A} + \log \mathbf{M}^{(g)}
\end{equation}
and the final output of this attention branch  \( \mathbf{Y}_g  \) could be obtained via
\begin{equation}
\mathbf{Y}_g = \text{softmax}(\mathbf{A}') \cdot \mathbf{V}
\end{equation}
As a result, each attention branch maintains broad receptive field coverage while significantly reducing computational overhead. In order to further adaptively integrate multi-granularity of sub-attentions, we incorporate lightweight granularity fusion network that adaptively assigns dynamic router weights for each branch, \ie
\begin{equation}
\mathbf{g} = \text{softmax}(\mathbf{W}(\text{layernorm}(\text{mean}(\mathbf{X}, 1)))+b).
\end{equation}
where $\mathbf{W}, \mathbf{b}$ are learnable parameters, the functions $\mathrm{mean}$ and $\mathrm{layernorm}$ denotes average pooling and layer normalization, respectively. Eventually, the final output of the proposed MoG attention is the weighted sum of all \( N_g \)  paralleled sub-attention branches that with different receptive fields, \ie

\begin{equation}
\mathbf{Y} = \sum_{e=1}^{N_g} g_g \cdot \mathbf{Y}_g
\end{equation}

As a result, the MoG attention attends to sparse positions, capturing and integrating various granularities of features and various range of dependencies across target scales in aerial images, eventually achieving scale-comprehensive representation learning.

\subsection{Comprehensive-to-Sensitive Decoder}
In order to better decode the scale-varying referring targets in aerial images, we design a two-stage comprehensive-to-sensitive decoding (CtS) strategy to decode in a coarse-to-fine manner. In first stage of scale-comprehensive decoder (SCD), the above mixture-of-granularity (MoG) attention are inserted into each transformer layer in a standard DETR-style decoder \cite{detr} for scale-comprehensive decoding. When dynamically integrating sub-attention branch with various receptive fields, the scale-comprehensive decoder effectively captures representations of various scales.

The second stage of scale-sensitive decoder (SSD) follows a standard DETR-style decoder \cite{detr} structure excepting its inputs of target queries and comprehensive representation embedding. As shown in Fig. \ref{fig:4}, its network structure contains sub-modules of multi-head self-attention (MHSA), multi-head cross-attention (MHCA), and feed forward network (FFN). It takes in coarse decoding results from the first stage to further conduct fine-grained scale-specific decoding. Besides, we hierarchically integrate multi-scale of representations from each block in scale-comprehensive encoder, thus obtaining comprehensive representations contain both intra-layer scale-comprehensive representations and inter-layer scale-comprehensive representations, serving as crucial guidance for fine-grained referring decoding. 

Eventually,\hspace{0.5em}the fine-grained query representation decoded from scale-sensitive decoder will be fed into the regression head to predict potential bounding-boxes.\hspace{0.5em}The Hungarian algorithm \cite{hungarian} is applied to efficiently search a bipartite matching between the prediction and annotation of the target.

\renewcommand{\thetable}{3}
\begin{table}[tbp] %h表示三线表在当前位置插入
\setlength{\abovecaptionskip}{0.05cm} %设置三线表标题与第一条线间距
\centering
\caption{Performance degradation from ground-view to aeriel-view datasets. Existing state-of-the-art ground referring detection methods undergo serious performance degradation on our aerial referring detection dataset (measured by the metric P@0.5).}
\setlength{\tabcolsep}{1mm}
\footnotesize
\begin{tabular*}{\linewidth}{@{\extracolsep{\fill}}l | c c c | c} % ★ 改这里
  \hline
  \hline
   \textbf{Methods} & \textbf{RefCOCO} & \textbf{RefCOCO+} & \textbf{RefCOCOg} & \textbf{RefAerial} \\
  \hline 
  TransVG \cite{transvg}& 80.32 & 63.50 & 67.11 & \textbf{13.91}  \\
  LLaVA-v1.5 \cite{llaval}& 86.52 & 80.17 & 82.32 & \textbf{25.12}  \\
  SegVG \cite{segvg} & 86.84 & 77.18 & 76.01 & \textbf{26.53}  \\
  SimVG \cite{simvg}  & 90.61 & 85.36 & 82.67 & \textbf{24.52} \\
  \hline
  \hline
\end{tabular*}
\label{tab:4}
\end{table}

\section{ Experiments}

\subsection{Implementation Details}
Regarding the main settings in the proposed scale-comprehensive and sensitive (SCS) framework. The backbone of the model is ResNet101, which is initialized with the publicly available DETR-R101-UNC checkpoint. Additionally, the pre-trained BERT module is incorporated to enhance the capability of language understanding. Following the standard protocol for referring detection in \cite{ma2024visual, zheng2025look,transvg}, we use precision (P@$\theta$) as the evaluation metric, where the prediction is deemed correct if its IoU with the ground-truth box is larger than threshold $\theta$. In detail, we utilize P@0.5, P@0.6, P@0.7, P@0.8, and mP (average pecision across these thresholds) except specially mentioned.

For hyperparameter settings, the initial learning rate for the entire model is set to $1 \times 10^{-4}$, while the BERT module, CNN backbone, and Transformer components in the visual encoder are fine-tuned with a lower learning rate of $1 \times 10^{-5}$ to mitigate overfitting and stabilize gradients. The model is trained for a total of 90 epochs, during which the backbone is frozen for the first 10 epochs to stabilize the convergence process in the early stages of training and improve the overall training performance.

\subsection{Comparison with State-of-the-Art Methods}
To establish a benchmark for the referring detection task in aerial images, we conduct performance comparison experiments based on several representative models, including 5 specialist models tailored for referring detection and 2 large multimodal models (LMMs). The specialist models include TransVG \cite{transvg}, SegVG \cite{segvg}, QRNet \cite{qrnet}, Dynamic-MDETR \cite{dynamic} and SimVG \cite{simvg}. As for the multimodal models, we evaluate the performance of Qwen-VL \cite{qwen} and LLaVA \cite{llaval}.

\textbf{Performance degradation:} 
In Table \ref{tab:4}, we compare the performance of the aforementioned mainstream referring detection methods on the RefCOCO series of ground-view datasets and our aerial image dataset RefAerial, using P@0.5 as the evaluation metric. The performance reported on RefAerial dataset is from the model re-trained/finetuned with its official code repository. Although these methods perform well on the RefCOCO series datasets (with  P@0.5 metric around 80\%), they all suffer a significant performance drop on the RefAerial dataset. The performance drop is primarily attributed to the critical characteristics of RefAerial, especially the characteristic that its target scales are very small and highly variable. This characteristic
not only hinders the ability of localizing targets in the visual domain, but also imposes hard challenge on language-to-vision alignment. These challenges collectively increase the difficulty of target localization, language understanding, and cross-modal alignment, and eventually constitute the core reasons of above performance degradation.

\textbf{Comparison on the RefAerial dataset:} 
In Table \ref{tab:3}, we present the performance comparison of several state-of-the-art referring detection methods on the RefAerial dataset, evaluated by the precision across different IoU thresholds. The results show that existing methods perform poorly at all localization thresholds. The RefAerial dataset contains challenging intrinsic characteristics in aerial referring detection, such as low object-to-scene ratio and numerous distractors, which expose the limitations of existing ground referring detection methods in modeling complex cross-modal contexts and accurately localizing densely packed small targets. These challenges make it difficult for conventional models to adapt to the intrinsic properties of aerial images. In contrast, the proposed SCS framework incorporates the MoG attention and the two-stage CtS decoding strategy, enabling more effective integration of multi-scale semantic information and contextual cues. As a result, it consistently outperforms existing methods across all evaluation metrics on the proposed RefAerial dataset and demonstrates stronger robustness and adaptability of referring detection in aerial-view scenarios.

 \textbf{Comparison on the RefCOCO series datasets:} To further examine the effectiveness of our scale-comprehensive and sensitive (SCS) framework on conventional ground referring detection datasets, we present the performance comparison of several state-of-the-art referring detection methods on the RefCOCO series datasets in Table \ref{tab:10}. The results show that our SCS even achevies some performance boost on conventional ground referring detection. However, the performance gains (+0.80,+0.57,+0.73) over the baseline (\ie, SegVG) on these ground datasets are much smaller than the gain (+1.62, Table \ref{tab:3}) on our aerial dataset, because the scale variety issue is much smaller in these ground referring detection datasets.

\renewcommand{\thetable}{4}
\begin{table}[tbp] %h表示三线表在当前位置插入
\setlength{\abovecaptionskip}{0.05cm} %设置三线表标题与第一条线间距
\small
\caption{Performance comparison with state-of-the-art methods on the RefAerial dataset.}
%表头文本加黑，但不加黑Table 1.字样，引入包即可：\usepackage[labelfont=bf]{caption}
 %设置三线表线条颜色：黑色
\setlength{\tabcolsep}{1mm}
\begin{tabular*}{\hsize}{@{\extracolsep{\fill}}l | c c c c | c} %{\hsize}使三线表自适应宽度，c表示文本居中
  \hline
  \hline
   \textbf{Methods} & \textbf{P@0.5} & \textbf{P@0.6} & \textbf{P@0.7} & \textbf{P@0.8} & \textbf{mP} \\
  \hline 
  TransVG \cite{transvg}* & 13.91& 10.17  & 7.31  & 3.23 & 8.65  \\
  Dyn. MDETR \cite{dynamic}*  & 21.25 & 16.87  &  10.24 & 4.56 & 13.23 \\
  SimVG \cite{simvg}*& 24.52 & 19.54  &  13.86& 4.93 & 15.71 \\
  LLaVA-v1.5 \cite{llaval}* & 25.12 & 18.63  & 10.97 & 4.62 & 14.83 \\
  Qwen-VL \cite{qwen}*& 26.23 & 19.62 & 12.57 & 5.01 & 15.85 \\
  SegVG \cite{segvg}* &  26.53 & 20.47 & 13.15& 5.41 & 16.39\\
  \hline
  \textbf{SCS (ours)}& \textbf{28.15} & \textbf{23.37} & \textbf{15.23} & \textbf{6.41} & \textbf{18.29} \\
  \hline
  \hline
\end{tabular*}
\noindent\textsuperscript{*} Our reimplementation from corresponding official repository.
\label{tab:3}
\end{table}

\renewcommand{\thetable}{5}
\begin{table}[h] %h表示三线表在当前位置插入
\setlength{\abovecaptionskip}{0.05cm} %设置三线表标题与第一条线间距
\centering
\caption{Performance comparison with state-of-the-art methods on conventional ground referring detection datasets (P@0.5).}
%表头文本加黑，但不加黑Table 1.字样，引入包即可：\usepackage[labelfont=bf]{caption}
 %设置三线表线条颜色：黑色
\setlength{\tabcolsep}{1mm}
\small %默认小一号字体
\begin{tabular*}{\hsize}{@{\extracolsep{\fill}}l | c c c } %{\hsize}使三线表自适应宽度，c表示文本居中
  \hline
  \hline
   \textbf{Methods} & \textbf{RefCOCO} & \textbf{RefCOCO+} & \textbf{RefCOCOg}
     \\
  \hline 
  TransVG \cite{transvg}& 80.32 & 63.50  & 67.11  \\
  QRNet \cite{qrnet}& 84.01 & 72.94 & 71.89\\
  LLaVA-v1.5 \cite{llaval}& 86.52 & 80.17 & 82.32  \\
  SegVG \cite{segvg} &  86.84 & 77.18 & 76.01\\
  \hline
  \textbf{SCS (ours)}& \textbf{87.64} \scriptsize(~$\uparrow$0.80) & \textbf{77.75} \scriptsize(~$\uparrow$0.57) & \textbf{76.74} \scriptsize(~$\uparrow$0.73)  \\
  \hline
  \hline
\end{tabular*}
\label{tab:10}
\end{table}

\renewcommand{\thetable}{6}
\begin{table}[!t]
\setlength{\abovecaptionskip}{0.1cm}
\centering
\small
\caption{Ablation study of our scale-comprehensive and sensitive (SCS) framework on the RefAerial dataset.}
\setlength{\tabcolsep}{3.5pt}
\renewcommand{\arraystretch}{1.2}
\begin{tabular}{l|l|l|c|c|c|c|c}
\hline
\hline
\textbf{Encoder} & \multicolumn{2}{l|}{\textbf{~  Decoder}} & \multicolumn{4}{c}{\textbf{Metrics}} \\
\cline{1-8}
  \textbf{~~~~SCE}& \textbf{SSD} & \textbf{SCD} & \textbf{P@0.5} & \textbf{P@0.6} & \textbf{P@0.7} & \textbf{P@0.8} & \textbf{mP} \\ 
\hline
~~~~~~× & ~~× & ~~× & 26.53 & 20.47 & 13.15 & 5.41 & 16.39 \\
~~~~~~\checkmark & ~~× & ~~× & 27.75 & 22.84 & 14.74 & 5.97 & 17.82 \\
~~~~~~× & ~~\checkmark & ~~× & 27.51 & 22.32 & 14.57 & 5.89 & 17.57 \\
~~~~~~× &~~× & ~~\checkmark & 26.83 & 21.82 & 13.21 & 5.47 & 16.83\\
~~~~~~\checkmark &~~× & ~~\checkmark & 27.83 & 22.52 & 14.21 & 5.97 & 17.63 \\
~~~~~~\checkmark &~~\checkmark & ~~× & 27.93 & 22.72 & 14.75 & 6.03 & 17.85 \\
~~~~~~\checkmark & ~~\checkmark & ~~\checkmark & \textbf{28.15} & \textbf{23.37} & \textbf{15.23} & \textbf{6.41} & \textbf{18.29} \\
\hline
\hline
\end{tabular}
\label{tab:2}
\end{table}

\subsection{Ablation Studies}
In order to analyze the effectiveness of each primary component in the proposed approach, we conduct systematic ablation studies on the RefAerial dataset. Starting from the baseline SegVG model, we progressively introduce the scale-comprehensive encoder (SCE), scale-comprehensive decoder (SCD), and scale-sensitive decoder (SSD) in various combinations. As illustrated in Table \ref{tab:2}, each module individually brings noticeable performance improvement. Specifically, the SCE and SCD are essential for effective extraction and integration of scale-comprehensive features, while SSD further refines the decoding of referring targets. When all three modules are included, the model achieves the best performance across all evaluation metrics, demonstrating the effectiveness and robustness of the overall design of our SCS framework in complex aerial scenarios.

\subsection{Impact of Granularity Number}
The granularity number in mixture-of-granularity (MoG) attention is one of key hyperparameters in the proposed SCS framework, thus we examine its impact on referring detection performance in Table \ref{tab:5}. We observe that increasing the attention granularities from 1 to 4 consistently improves the detection performance since incorporating more granularities are capable to capture more comprehensive scale variations. However, when the granularity number exceeds 4, the performance begins to gradually degrade. Therefore, based on extensive prior empirical observations, we select four attention granularities as the most optimal and stable configuration.

%components driving performance improvement. Leveraging multiple attention branches with different granularities and receptive fields, our MoG effectively captures scale-comprehensive fine-grained details and coarse-grained context. Compared to standard attention with uniform receptive fields, the proposed MoG offers superior scale-awareness, enhancing scale-diverse referring target perception within and across aerial images with low but diverse object-to-scene ratios.\hspace{0.5em}As shown in Table \ref{tab:5}, we observe that increasing the number of attention granularities from 1 to 4 consistently improves segmentation performance, indicating that incorporating granularity indeed captures more comprehensive scale variations. However, when the number of experts exceeds 4, the performance begins to saturate, suggesting four attention granularities strike a favorable balance between complexity and effectiveness.

\renewcommand{\thetable}{7}
\begin{table}[!t]
\centering
\setlength{\abovecaptionskip}{0.1cm} %设置三线表标题与第一条线间距
\captionsetup{labelsep=period}
\caption{Impact of the granularity number in the MoG attention.}
%表头文本加黑，但不加黑Table 1.字样，引入包即可：\usepackage[labelfont=bf]{caption}
\setlength{\tabcolsep}{1mm}
\begin{tabular*}{\hsize}{@{\extracolsep{\fill}}c | c c c c | c} %{\hsize}使三线表自适应宽度，c表示文本居中
  \hline
   \hline
   \textbf{Granularity}  & \textbf{P@0.5} & \textbf{P@0.6} & \textbf{P@0.7} & \textbf{P@0.8} & \textbf{mP} \\
  \hline 
   1  & 25.52  & 20.96  &  13.45 & 5.17 & 16.27\\
   2  & 26.07 & 21.23 & 14.32 & 5.65 & 16.81 \\
   3  &27.51 & 22.13 & 14.75  & 6.12 & 17.62 \\
   4  & \textbf{28.15} & \textbf{23.37} & \textbf{15.23} & \textbf{6.41} & \textbf{18.29}\\
   5  & 28.05 & 23.17 &15.01  & 6.15 & 18.09\\
   6  & 27.82 & 22.23 & 14.67 & 5.93 & 17.66\\
    \hline
   \hline
\end{tabular*}
\label{tab:5}
\end{table}

\subsection{Visualization Analysis}
\begin{figure}[h]  % 使用figure环境可以为图片添加标题，并自动管理图片位置
    \centering
    \includegraphics[page=1, width=0.8\linewidth, keepaspectratio]{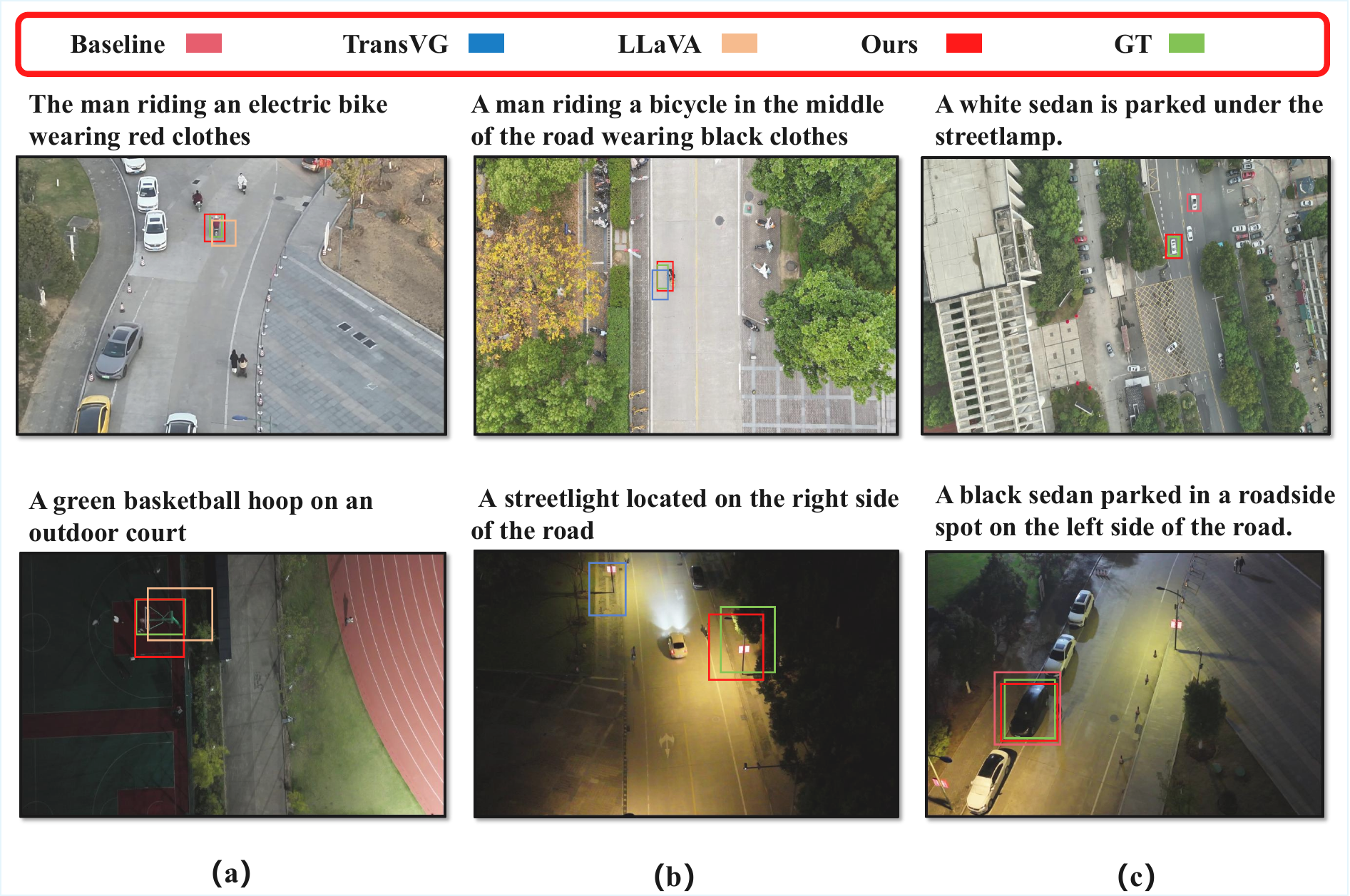}
    \caption{(a) Visualization comparison with methods with large models (such as LLaVA). (b) Visualization comparison with methods without large models (such as TransVG). (c) Visualization comparison with baseline method (\ie~SegVG).}  % 图片标题
    %Visualization of referring detection results on RefAerial from different approaches. 
    \label{fig:5}  % 可选的标签，便于在文档中引用
\end{figure}

Fig. \ref{fig:5} provides an intuitive effectiveness evaluation of the proposed approach. Compared to
classical methods tailored for ground-view referring detection (such as LLaVA \cite{llaval} and TransVG \cite{transvg}), our SCS framework exhibits stronger referring detection performance in various challenging aerial scenarios regarding both recognition and localization accuracy. Fig.\hspace{0.5em}\ref{fig:5} shows that our SCS achieves superior localization accuracy and boundary alignment compared to the baseline SegVG \cite{segvg}. Specifically, the baseline model fails to identify targets with large scale variation or fine-grained referring expressions. When enhanced with the proposed scale-comprehensive and sensitive (SCS) framework, it produces more accurate prediction. These visualizations intuitively validates the effectiveness and superiority the proposed method in aerial referring detection. %These results validate that the proposed modules bring significant performance improvements in referring detection tasks.

\section{Conclusion}

In this work, we propose a large-scale dataset tailored for referring detection in aerial images, which distinguishes from conventional ground referring detection datasets by 4 characteristics. For efficient referring annotation, we design a semi-automated referring annotation engine REA-Engine that effectively exploiting the multimodal foundation models with human feedback. Additionally, we propose a SCS framework that integrates the MoG attention and a CtS decoding strategy to improve aerial referring detection. In future work, we will extend the SCS framework to address more core challenges in aerial referring detection and expand RefAerial for additional referring understanding tasks, including referring segmentation and counting.

{
    \small
    \bibliographystyle{ieeenat_fullname}
    \bibliography{main}
}


\end{document}